\documentclass[11pt,a4paper]{article}

\usepackage[utf8]{inputenc}
\usepackage[T1]{fontenc}
\usepackage{lmodern}
\usepackage{amsmath, amssymb}
\usepackage{graphicx}
\usepackage{hyperref}
\usepackage{longtable}
\usepackage{geometry}
\usepackage{url} % Añade esto en el preámbulo

\title{Enhancing Forex Forecasting Accuracy: The Impact of Hybrid Variable Sets in Cognitive Algorithmic Trading Systems}

\author{
  Juan C. King\\
  Centro de Investigación Operativa, Universidad Miguel Hernández, 03202 Elche, Spain\\
  \texttt{juan.king@goumh.umh.es}
  \and
  José M. Amigó\\
  Centro de Investigación Operativa, Universidad Miguel Hernández, 03202 Elche, Spain\\
  \texttt{jm.amigo@umh.es}
}

\date{2025}

\begin{document}

\maketitle

% Abstract (Do not insert blank lines, i.e. \\) 
\abstract{This paper presents the implementation of an advanced artificial intelli\-gence-based algorithmic trading system specifically designed for the EUR-USD pair within the complex and high-frequency environment of the Forex market. The methodological approach centers on integrating a holistic set of input features: key fundamental macroeconomic variables (e.g., Gross Domestic Product, Unemployment Rate) collected from both the Euro Zone and the American Zone, alongside a comprehensive suite of technical variables (including various indicators, oscillators, Fibonacci levels, and price divergences). The performance of the resulting algorithm is subjected to a rigorous evaluation process. This assessment utilizes both standard Machine Learning metrics to quantify model accuracy and backtesting simulations across historical data to evaluate trading profitability and risk. Crucially, the study concludes with a detailed comparative analysis designed to determine which class of input features—fundamental or technical—provides the system with the greater and more reliable predictive capacity for generating profitable trading signals.}

\section{Introduction}

\label{s:intro}

\subsection{Background}

The question whether algorithmic trading systems (ATS) can improve human trading in terms of effectiveness is eliciting an increasingly relevant debate among traders and investors, as well as quantitative studies that address this issue through numerical testing [\cite{demirtacs2020algorithmic}]. 

In recent years, the discussion regarding whether algorithmic trading systems (ATS) can surpass human traders in terms of efficiency, consistency, and adaptability has gained significant traction in both academic and professional circles. Empirical evidence indicates that algorithmic strategies tend to exhibit superior performance in volatile or declining markets, whereas human-managed funds may retain a relative advantage during upward market trends due to behavioral and intuitive factors [\cite{anuar2025comparative}]. Moreover, large-scale behavioral studies reveal that algorithms largely eliminate well-known cognitive biases such as the disposition effect that continue to affect human traders [\cite{liaudinskas2022human}]. Complementary research has also emphasized the growing integration of artificial intelligence and machine learning methods in modern ATS, which enhances predictive accuracy and execution speed [\cite{cohen2022algorithmic}]. Nonetheless, experimental findings suggest that algorithmic trading may still be constrained by design limitations, challenging the notion of its absolute superiority over human decision-making [\cite{jacob2024algorithmic}]. These findings collectively indicate that algorithmic and human trading approaches might be best viewed as complementary, each offering unique strengths under different market conditions.

Thus, the advantages and disadvantages of algorithmic trading were analyzed in [\cite{reznik2018high}]. Among the advantages, the authors point out that algorithmic trading is not affected by emotions (fear and greed), it has the possibility of back-testing, it is very fast and allows operations to be diversified. On the other hand, technical errors, optimization errors and excessive optimization are identified as disadvantages. The authors conclude that human trading cannot be replaced by algorithmic trading.

Before introducing the concept of a cognitive Algorithmic Trading System (cognitive ATS), it is essential to clarify what a cognitive system is. In general terms, a cognitive system refers to an artificial intelligence architecture capable of perceiving its environment, learning from experience, reasoning about complex patterns, and adapting its actions to achieve specific goals autonomously. Unlike conventional algorithmic systems that follow predefined rules, cognitive systems integrate perception, memory, learning, and decision-making processes inspired by human cognition [\cite{wang2022cognitive}]. When applied to trading, a cognitive ATS would not only execute predefined strategies but also interpret market signals, evaluate uncertainty, and modify its behavior dynamically based on contextual information and prior outcomes.

Reference [\cite{martin2019cognitive}] provides a theoretical framework for an approach to what a cognitive ATS would require. The author talks about different types of data sets such as fundamental data macro parameters like EBITDA, ROI (Return of Investment), etc), market data (price, volume, dividends), analytic data (recommendations, market sentiments), and other data (images, Google searching, social networks).

Another approach to cognitive ATS  uses a decision support system based on an extended approximate cognitive network [\cite{li2020intelligent}], which is a method that combines elements of artificial neural networks and the theory of approximate approximation for analysis and decision making in financial markets. Let us remind that artificial neural networks are computational models that consist of layers of interconnected nodes that process information and learn patterns from historical data sets [\cite{lecun2015deep}]. In the context of trading, artificial neural networks are used to analyze market data, identify patterns and make predictions about future price behavior. As for the theory of approximate approximation, it focuses on approximation and estimation rather than exact precision [\cite{mazya2018approximate}]. In algorithmic trading, this means that, instead of trying to predict market behavior with absolute certainty, the aim is to obtain estimates that are as accurate as possible to make informed decisions on financial operations. The results of this approach reported in [\cite{li2020intelligent}] are positive.

However, in relation to the debate on algorithmic trading versus human trading, there is a question that has not been answered yet in the literature, namely: Is it possible to perform an analysis by a human agent that is not possible computationally? This is a question that we try to discuss in this paper from a scientific point of view.

%%%%%%%%%%%%%%%%%%%%%%%%%%%%%%%%%%%%%%%%%%%%%%%%%%%%%%

\subsection{Objectives}

{\normalsize \label{s:objectives} }

The objective of this work is the implementation by recurrent neural networks of an ATS that, based on both fundamental and technical data, achieves a statistical advantage large enough to qualify as a cognitive algorithmic trading system. Our approach is to introduce into a predictive model at least the same information that a human trader can analyze when making decisions about purchase and sale operations, and then verify if the algorithmic trading system is able of outperform it.

To develop such a system, the EUR-USD currency pair will be employed, as commonly done in similar studies [\cite{fischer2018deep}]. The dataset will include both fundamental macroeconomic variables and technical features derived from price data, such as indicators, oscillators, support and resistance levels, and divergences. These variables will be used to train models based on Long Short-Term Memory (LSTM) recurrent neural networks. LSTM networks are particularly suitable for this application due to their ability to retain historical information over long sequences, thereby improving predictive performance in time series analysis.

The approach involves incorporating all relevant variables into a single predictive algorithm. Price-related variables are represented as high-frequency time series, whereas fundamental macroeconomic indicators are released only monthly or quarterly, making it challenging to align all variables in a single dataset. To address this, the datasets were constructed by assigning to each day the most recently published value of each macroeconomic indicator. Additionally, a feature capturing the number of days since the latest release was included, allowing the model to weight information according to its recency during the training process.

Once we have analyzed which groups of data sources have the best performance based on the metrics obtained, it is necessary to create the best model with the set of variables that have the best performance. And to that model apply a simulation with data that the model has never seen before in buying and selling operations to check if the statistical advantage is good enough for the system to be profitable.

%%%%%%%%%%%%%%%%%%%%%%%%%%%%%%%%%%%%%%%%%%%%%%%%%%

\subsection{Contents}

This paper is organized as follows.

%\begin{enumerate}
\begin{itemize}

\item \textbf{Section 2} - "Variables" - details the variables used in the analysis, including fundamental variables and technical variables such as indicators, oscillators, supports and resistances, Fibonacci levels, and convergences/divergences. Target, defines the prediction target of the model (e.g., price movement, returns).

\item \textbf{Section 3} - "Methodology" - describes the model configurations, dataset construction process, training and evaluation procedures, and presents the results.

\item \textbf{Section 4} - "Trading Simulation and Performance Evaluation" - explains how the model can be practically applied in trading strategies or portfolio management.

\item \textbf{Section 5} - "Discussion" - interprets the results, compares findings with existing literature, and discusses limitations of the study.

\item \textbf{Section 6} - "Conclusion" - summarizes key findings, discusses their implications, and suggests directions for future work.

%\end{enumerate}
\end{itemize}

%%%%%%%%%%%%%%%%%%%%%%%%%%%%%%%%%%%%%%%%%%%%%%%%%%%%%%

\section{Variables}

\label{s:methods}

In this section, we describe the fundamental and technical variables used in this article. This is the information that will be passed to our predictive models, which consist of Long Short-Term Memory (LSTM) networks. Furthermore, we also describe a binary variable called Target; its value at the current time gives a projection of the price direction with the desired prediction horizon.

%%%%%%%%%%%%%%%%%%%%%%%%%%%%%%%%%%%%%%%%%%%%%%%%%%

\subsection{Fundamental Variables}
\label{s:variables.1}

The fundamental variables consist of eight datasets with macroeconomic information of the United States (US), retrieved from the Alpha Vantage API, and another eight datasets with the same information of the Euro Area (EA), retrieved from the Eurostat API, totaling 16 macro-economical variables. The temporal range of the dataset spans from January 1, 2012, to March 1, 2023. 

The macroeconomic data of the US and the EA that we have included as fundamental variables in the present study are the following [\cite{chen2010news}]. 

\begin{enumerate}
\item HIPC - Inflation Rate (i.e., obtained from the Harmonized Index of Consumer Prices ---HICP).

\item HIPC - Contributions to the Annual Inflation (in Percentage Points).
    
\item Annual Unemployment Rate.
    
\item Quarterly Unemployment Rate, Seasonally Adjusted.
    
\item Net External Debt - Quarterly Data, \% of GDP.
    
\item General Government Gross Debt (EDP Concept), Consolidated - Annual Data (i.e., according to the Excessive Deficit Procedure ---EDP).
    
\item Government Consolidated Gross Debt by Components - Annual Data.
    
\item General Government Gross Debt (EDP Concept), Consolidated - Quarterly Data.
    
\end{enumerate}

Due to the differences in the publication frequency of the data 1-8 (annually, quarterly, monthly), for each variable we kept track of the number of days since the last publication date, so that the LSTM net could establish a relationship between each variable and the number of days since the corresponding data publication.

%%%%%%%%%%%%%%%%%%%%%%%%%%%%%%%%%%%%%%%%%%%%%%%%%%

\subsection{Technical variables}

The technical variables used in the predictive models in Section \ref{s:methods.1.1} are of four types: (i) indicators and oscillators, (ii) supports and resistances, (iii) Fibonacci levels, and (iv) convergences and divergences. We detail them in the following sections.

%%%%%%%%%%%%%%%%%%%%%%%%%%%%%%%%%%%%%%%%%%%%%%%%%%%%%%

\subsubsection{Indicators and Oscillators}
\label{s:variables.2}

Indicators and oscillators [\cite{king2024blockchain}] are based on the daily-closing price of the Euro-Dollar pair, from December 1, 2003 to February 22, 2024.

The following list specifies the 11 indicators and oscillators considered here.

\begin{enumerate}

\item Simple Moving Average (SMA) [\cite{thompson1947use}]. The formula is given by
\begin{equation*}
\text{SMA}_{n} = \frac{1}{n} \sum_{i=1}^{n} \text{Close}_{i}
\end{equation*}
where $n$ is the size of the time window (number of periods in the window), and $\text{Close}_{i}$ is the closing price at the $i$-th day of the window. $\text{SMA}_{n}$ was calculated for $n=20, 55$.

\item Exponential Moving Average (EMA) [\cite{klinker2011exponential}]. The formula is given recurrently by
\begin{equation*}
\text{EMA}_{n} = \text{EMA}_{n-1} + \frac{2}{n + 1} \times (\text{Close}_{n} - \text{EMA}_{n-1})
\end{equation*}

where $n$ is the size of the time window, and $\text{Close}_{n}$ is the closing price at the last day of the window. $\text{EMA}_{n}$ was calculated for $n=20, 55, 200$.

\item Bollinger Bands [\cite{bollinger2002bollinger}] are envelopes plotted at a standard deviation level above and below a simple moving average of the price. They are defined by the formulas
\begin{align*}
\text{Lower band:} \;\; \text{BBL}_{n,k} &= \text{SMA}_{n}(\text{Close}) - k \times \text{StdDev}_{n}(\text{Close}) \\
\text{Middle band:} \; \text{BBM}_{n,k} &= \text{SMA}_{n}(\text{Close}) \\
\text{Upper band:} \;\; \text{BBU}_{n,k} &= \text{SMA}_{n}(\text{Close}) + k \times \text{StdDev}_{n}(\text{Close}) \\
\text{Band width:} \;\; \text{BBB}_{n,k} &= \frac{\text{BBU}_{n,k} - \text{BBL}_{n,k}}{\text{BBM}_{n,k}} \\
\text{Band percent:} \;\; \text{BBP}_{n,k} &= \frac{\text{Close} - \text{BBL}_{n,k}}{\text{BBU}_{n,k} - \text{BBL}_{n,k}}
\end{align*}

where $n$ is the size of the time window, $k$ is the number of standard deviations of the data in the window ($\text{StdDev}$), $\text{Close}$ is the closing price, $\text{StdDev}_{n}(\text{Close})$ is the standard deviation of the closing prices over the window, and $\text{SMA}_{n}(\text{Close})$ is the simple moving average of the closing prices over the window. 

The parameter $n$ determines the sensitivity of the bands to changes in prices. The parameter $k$ determines the distance between the bands and the moving average. We used the standard parametric values: $n=20$ and $k=2$.

\item Ichimoku Cloud [\cite{elliott2007ichimoku}] is defined by the formulas
\begin{align*}
\text{ITS}_{n} &= \frac{{\text{High} + \text{Low}}}{2} \, \;(\text{calculated over the last $n$ periods}) \\ 
\text{IKS}_{m} &= \frac{{\text{High} + \text{Low}}}{2} \, \; \text{(calculated over the last $m$ periods)} \\
\text{ISA}_{p} &= \frac{{\text{ITS} + \text{IKS}}}{2} \, \; 
\text{(calculated over the last $p$ periods, plotted $\frac{p}{2}$ periods ahead)} \\
\text{ISB}_{p} &= \frac{{\text{High} + \text{Low}}}{2} \, \; (\text{calculated over the last $p$ periods, plotted $\frac{p}{2}$ periods ahead}) \\
\text{CS}_{m} &= \text{Closing Price} \, \;(\text{plotted $m$ periods back)}
\end{align*}

The "Tenkan Sen" line $\text{ITS}_{n}$ is called the conversion line; the "Kijun Sen" line $\text{IKS}_{m}$ is called the base line; the "Senkou Span A" line $\text{ISA}_{p}$ is called the leading span A; and the "Senkou Span B" line $\text{ISB}_{n}$ is called the leading span B; and the "Chikou Span" line $\text{CS}_{m}$ is called the lagging span. High and Low refer to the highest and lowest prices in the corresponding time window. We used the standard parametric values: $n = 9$, $m = 26$, and $p = 52$.

\item Relative Strength Index (RSI) [\cite{wu2015technical}] is defined as
\begin{equation*}
\text{RSI}_{n} = 100 - \frac{100}{1 + \text{RS}}
\end{equation*}
where
\begin{equation*}
\text{RS} = \frac{\text{Average Gain}_{n}}{\text{Average Loss}_{n}}.
\end{equation*}

Here, $\text{Average Gain}_{n}$ is the average gain over the last $n$ days, and $\text{Average Loss}_{n}$ is the average loss over the last $n$ days. RSI was calculated for $n = 6, 12, 14, 24$.

\item Moving Average Convergence Divergence (MACD) [\cite{appel1985moving}] is defined by the formulas
\begin{align*}
\text{MACD}_{n,m} &= \text{EMA}_{n}(\text{Close}) - \text{EMA}_{m}(\text{Close}) \\
\text{MACDh}_{n,m,p} &= \text{MACD}_{n,m} - \text{EMA}_{p}(\text{MACD}_{n,m}) \\
\text{MACDs}_{n,m,p} &= \text{EMA}_{p}(\text{MACD}_{n,m})
\end{align*}

where $n$ is the size of the time window for the shorter EMA, $m$ is size of the time window for the longer EMA ($m>n$), $p$ is the number of periods of the EMA signal line, $\text{EMA}_{n}(\text{Close})$ is the exponential moving average of the closing prices in the window of size $n$, $\text{MACDh}_{n,m,p}$ is the MACD histogram, and $\text{MACDs}_{n,m,p}$ is the MACD signal line. We used the standard parametric values: $n= 12$, $m= 26$ and $p=9$.

\item Average Directional Index (ADX) [\cite{salkar2021algorithmic}] is calculated by the loop
\[ \text{ADX}_{t} = \frac{\text{ADX}_{t-1} \times (n-1) + \text{DX}_{t}}{n} \] 
where $n$ is the time window, $t=1,...,n$, 
\[ \text{DX}_{t} = \frac{\left| \text{DI}^{+}_{t} - \text{DI}^{-}_{t}\right|}{\left| \text{DI}^{+}_{t} + \text{DI}^{-}_{t} \right|} \] 
and
\[ \text{DI}^{+}_{t} = \text{High}_{t} - \text{High}_{t-1} \] 
\[ \text{DI}^{-}_{t} = \text{Low}_{t-1} - \text{Low}_{t} \] 
where $\text{High}_{t}$ (resp., $\text{Low}_{t}$) is the highest (resp., lowest) price at period $t$. As usual, we set $n=14$. 

\item Williams (\%R) [\cite{steele2015technical}] is given by
\begin{equation*}
\text{Williams \%R}_{n} = \frac{{\text{High}_{n} - \text{Close}}}{{\text{High}_{n} - \text{Low}_{n}}} \times 100\%
\end{equation*}

where $\text{High}_{n}$ is the highest price over the last $n$ periods, $\text{Low}_{n}$ is the lowest price over the last $n$ periods, and $\text{Close}$ is the closing price (at the last period). We used the standard value $n=14$.

\item Average True Range (ATR) [\cite{panapongpakorn2019short}] is a price volatility indicator defined as
\begin{equation*}
\text{ATR}_{n} = \frac{1}{n} \sum_{i=1}^{n} \max(\text{High}_{i} - \text{Low}_{i},|\text{High}_{i} - \text{Close}_{i-1}|, |\text{Low}_{i} - \text{Close}_{i-1}|)
\end{equation*}

where $n$ is the size of the time window, $\text{High}_{i}$ is the highest price in the periods $1,2,...,i$, $\text{Low}_{i}$ is the lowest price in the periods $1,2,...,i$, and $\text{Close}_{i-1}$ is the closing price at period $i-1$. We used the standard value $n=14$.

\item Stochastic Oscillator (KDJ) [\cite{wu2015technical}] is an indicator that measures the current price of an asset in relation to its range over a time interval. It is defined as
\begin{align*}
\%\text{K} &= \frac{{\text{Close} - \text{Min Low}_{n}}}{{\text{Max High}_{n} - \text{Min Low}_{n}}} \times 100\\
\%\text{D} &= \text{SMA}_{n}(\%\text{K})\\
\text{J} &= 3 \times \%\text{K} - 2 \times \%\text{D}
\end{align*}

where $\text{Close}$ is the current closing price, $\text{Min Low}_{n}$ is the asset lowest price over the last $n$ periods, and $\text{Max High}_{n}$ is the highest price over the same $n$ periods.

We used the standard parametric values: calculation time interval for \%\text{K} = 14, and
\%\text{D} smoothing window size = 3.

\item Squeeze Momentum (SQZ) [\cite{alostad2017directional}] is a volatility indicator defined as

\begin{equation*}
\text{SQZ}_{n,m,p,q} = \frac{\text{SMA}(\text{Close}_{n}) - \text{SMA}(\text{Close}_{m})}{\text{SMA}(\text{Close}_{p}) \times q}
\end{equation*}
where $n$ is the size of the time window for the shorter mean (usually ranging from 5 to 20), $m$ is the size of the time window for the longer mean (usually between 20 and 50), $p$ is the size of the time window for the comparison mean, $q$ is a multiplier (typically 2 or 3), and $\text{SMA}(\text{Close}_{n})$ is the simple average of the closing prices in the window of size $n$. 

We used the following parametric values: $n = 20$, $m = 50$, $p = 200$, and $q = 2$.

\end{enumerate}

%%%%%%%%%%%%%%%%%%%%%%%%%%%%%%%%%%%%%%%%%%%%%%%%%

\subsection{Computation of Support and Resistance Levels}

\label{s:variables.2.2}

Mathematically, support and resistance zones can be defined as price regions where local maxima and minima tend to cluster [\cite{osler2000support}]. 
In our approach, we estimate two support levels and two resistance levels for each observation by statistically grouping recent price extrema. 
The computation proceeds as follows.

\begin{enumerate}
    \item \textbf{Window segmentation.}  
    From the last 200 daily observations, we define ten non-overlapping windows of 20 days each, denoted as \( w_i \), with \( i = 1, 2, \dots, 10 \).
    \[
    w_i = \{ P_t \mid t \in [20(i-1)+1, 20i] \}.
    \]

    \item \textbf{Extraction of local extrema.}  
    For each window \( w_i \), we identify the maximum and minimum of both closing prices and intraday highs and lows:
    \[
    M_i = \{ \max(\text{High}_{w_i}), \max(\text{Close}_{w_i}) \}, \quad 
    m_i = \{ \min(\text{Low}_{w_i}), \min(\text{Close}_{w_i}) \}.
    \]
    All extrema are collected into a single sorted list:
    \[
    L = \text{sort}\left( \bigcup_{i=1}^{10} (M_i \cup m_i) \right).
    \]

    \item \textbf{Clustering of nearby extrema.}  
    The list \( L = \{ x_1, x_2, \dots, x_n \} \) is passed to a custom function \texttt{grouper}, which aggregates consecutive elements that are close in value.  
    Two elements \( x_i \) and \( x_{i+1} \) belong to the same group if their difference is smaller than a proximity threshold \( \delta \):
    \[
    |x_{i+1} - x_i| < \delta,
    \]
    where \( \delta = \alpha P_t \), with \( \alpha = 0.04 \) representing a $4\%$ tolerance relative to the current price \( P_t \).

    Formally, the function returns a list of subsets
    \[
    \mathcal{G} = \{ G_1, G_2, \dots, G_k \},
    \]
    where
    \[
    G_j = \{ x_i \in L \mid |x_{i+1} - x_i| < \delta \text{ for all consecutive } x_i, x_{i+1} \in G_j \}.
    \]
    Each group \( G_j \) therefore represents a region where price extrema are concentrated.

    \item \textbf{Computation of representative levels.}  
    For each group \( G_j \), a representative price level is obtained as the arithmetic mean:
    \[
    \bar{x}_j = \frac{1}{|G_j|} \sum_{x_i \in G_j} x_i.
    \]
    The resulting set of mean values
    \[
    S = \{ \bar{x}_1, \bar{x}_2, \dots, \bar{x}_k \}
    \]
    defines the potential support and resistance zones.

    \item \textbf{Identification of active levels.}  
    Given the current price \( P_t \), the two nearest levels below it are defined as supports:
    \[
    \text{Support}_1 = \max\{ \bar{x}_j \in S \mid \bar{x}_j < P_t \}, \quad
    \text{Support}_2 = \max\{ \bar{x}_j \in S \mid \bar{x}_j < \text{Support}_1 \},
    \]
    and the two nearest levels above it as resistances:
    \[
    \text{Resistance}_1 = \min\{ \bar{x}_j \in S \mid \bar{x}_j > P_t \}, \quad
    \text{Resistance}_2 = \min\{ \bar{x}_j \in S \mid \bar{x}_j > \text{Resistance}_1 \}.
    \]

    \item \textbf{Integration into the dataset.}  
    These four values—second-last resistance, last resistance, last support, and second-last support—were stored as new variables in the dataset for further analysis and model training.
\end{enumerate}

%%%%%%%%%%%%%%%%%%%%%%%%%%%%%%%%%%%%%%%

\label{s:variables.2.3}

\subsubsection{Fibonacci Retracement Levels}

The set of technical variables is further expanded by incorporating two additional support and two resistance levels derived from Fibonacci retracement principles.

Fibonacci retracement levels are a prominent tool in technical analysis, derived from key ratios of the Fibonacci sequence. The sequence, defined recursively as \( F_n = F_{n-1} + F_{n-2} \) with seed values \( F_0 = 0 \) and \( F_1 = 1 \), gives rise to ratios that are believed to signify potential levels of market significance [\cite{murphy1999technical}]. In financial markets, these levels are employed to identify probable areas of support and resistance during price retracements within a larger trend.

The primary retracement levels used in this study are 23.6\%, 38.2\%, 50\%, 61.8\%, and 78.6\%. These percentages correspond to key Fibonacci ratios; notably, 38.2\% and 61.8\% approximate \( 1 - \phi^{-1} \) and \( \phi^{-1} \), respectively, where \( \phi \) (the golden ratio) is approximately 1.618. While the 50\% level is not a pure Fibonacci ratio, it is conventionally included due to its widespread use among practitioners. The predictive relevance of these levels often stems from their self-fulfilling nature, as their collective recognition by market participants can influence trading decisions and price action [\cite{shaker2018predictive}].

The methodology for calculating these levels mirrors the procedure outlined in Section \ref{s:variables.2.3}. For each trading day in the dataset, the maximum (High) and minimum (Low) prices over the preceding 200-day rolling window were identified. This price range was then used to compute the standard suite of Fibonacci levels, including the 127.2\% and 161.8\% extension levels. Subsequently, the algorithm determined the two closest support and two closest resistance levels relative to the current closing price. These four values---denoting the immediate Fibonacci-based supports and resistances---were appended to the dataset as new technical variables.

%%%%%%%%%%%%%%%%%%%%%%%%%%%%%%%%%%%%%%%

\subsubsection{Convergences and Divergences}
\label{s:variables.2.4}

This section details the methodology for identifying convergence and divergence between the price action and the Squeeze Momentum indicator (see Section \ref{s:variables.2}, indicator 11). These concepts are fundamental in technical analysis for assessing the strength of a trend and anticipating potential reversals.

The Squeeze Momentum indicator, which synthesizes concepts from Bollinger Bands (Section \ref{s:variables.2}, indicator 3) and the Momentum oscillator [\cite{blau1995momentum}], is designed to identify periods of market compression (low volatility) and subsequent expansion (high volatility). Its value reflects the underlying momentum once the market exits a low-volatility state.

In this context, convergence is defined as a scenario where the prevailing trends in the price and the indicator are aligned. Conversely, divergence is identified when the trends in the price and the indicator are in opposition. Classically, convergence suggests a continuation of the current trend, while divergence signals a potential weakening of the trend and an increased probability of a reversal or retracement [\cite{blau1995momentum, kirkpatrick2010technical}].

The algorithmic procedure for detecting these patterns at a given time \( t \) is as follows:

\begin{enumerate}
    \item \textbf{Identify Peaks and Troughs:} For both the price series and the Squeeze Momentum indicator, identify the two most recent local highs and the two most recent local lows within a rolling window of the past 40 days.

    \item \textbf{Construct Trend Lines:}
    \begin{itemize}
        \item Calculate the linear trend line connecting the two identified price highs.
        \item Calculate the linear trend line connecting the two identified price lows.
        \item Calculate the linear trend line connecting the two corresponding indicator highs.
        \item Calculate the linear trend line connecting the two corresponding indicator lows.
    \end{itemize}

    \item \textbf{Calculate Slopes:} Compute the slopes of the four trend lines:
    \begin{itemize}
        \item \( m_{\text{price}}^{\text{high}} \): Slope of the price highs trend line.
        \item \( m_{\text{price}}^{\text{low}} \): Slope of the price lows trend line.
        \item \( m_{\text{ind}}^{\text{high}} \): Slope of the indicator highs trend line.
        \item \( m_{\text{ind}}^{\text{low}} \): Slope of the indicator lows trend line.
    \end{itemize}

    \item \textbf{Classify the Regime:} The market state is classified separately for highs and lows based on the sign of the product of the slopes. A positive product indicates convergence (aligned trends), while a negative product indicates divergence (opposing trends).
    \begin{itemize}
        \item \textbf{For Highs:} The state \( S_{\text{high}} \) is given by the sign of \( m_{\text{price}}^{\text{high}} \cdot m_{\text{ind}}^{\text{high}} \).
        \item \textbf{For Lows:} The state \( S_{\text{low}} \) is given by the sign of \( m_{\text{price}}^{\text{low}} \cdot m_{\text{ind}}^{\text{low}} \).
    \end{itemize}
    A value of \( S > 0 \) denotes convergence, and \( S < 0 \) denotes divergence. These two binary states, \( S_{\text{high}} \) and \( S_{\text{low}} \), are added to the dataset as two new technical variables for time \( t \).
\end{enumerate}

%%%%%%%%%%%%%%%%%%%%%%%%%%%%%%%%%%%%%%%%%%%%%%%%%%

\subsection{Target Variable}
\label{s:variables.3}

In line with the methodology established in our previous research [\cite{king2025integration}], the target variable for this study is a binary indicator reflecting future market direction. This formulation allows for a clear assessment of whether the proposed model achieves a statistical forecasting advantage.

Consider a time series of daily closing prices, denoted as \( \{P(t)\}_{t=1}^{T} \). For a given observation at time \( n \) and a predefined prediction horizon \( h = 10 \) days (such that \( n + h \leq T \)), we define the \textit{directional index} \( D(n, h) \) as follows:
\begin{equation}
D(n, h) = \frac{1}{2}\left( \max_{i \in [n+1, n+h]} P(i) - P(n) \right) + \frac{1}{2}\left( \min_{i \in [n+1, n+h]} P(i) - P(n) \right) + \frac{1}{2}\left( P(n+h) - P(n) \right)
\label{eq:directional_index}
\end{equation}
where \( P(n) \) is the closing price at time \( n \), \( \max_{i \in [n+1, n+h]} P(i) \) is the maximum closing price in the forward-looking window, and \( \min_{i \in [n+1, n+h]} P(i) \) is the corresponding minimum closing price.

This composite measure \( D(n, h) \) assigns equal weight to three critical price paths: the maximum attainable profit, the maximum potential loss, and the final price movement over the horizon. This design is particularly relevant for automated trading systems, where positions may be closed prior to the horizon's end due to profit-taking or stop-loss mechanisms.

The binary target variable \( Y(n) \) is then derived by assessing the sign of the directional index relative to the initial price:
\begin{equation}
Y(n) = \begin{cases} 
1 & \text{if } D(n, h) > 0 \\
0 & \text{if } D(n, h) \leq 0 
\end{cases}
\label{eq:binary_target}
\end{equation}

While a continuous measure could provide more granularity, the binary classification suffices for the primary objective of this work: to determine the model's capability to predict the direction of price movement (upward: \( Y(n)=1 \), or downward: \( Y(n)=0 \)).

%%%%%%%%%%%%%%%%%%%%%%%%%%%%%%%%%%%%%%%%%%%%%%

\section{Methodology}

Following the data collection and preprocessing stages, we conducted a comparative analysis of various Long Short-Term Memory (LSTM) network architectures. The experimental framework was designed to evaluate the predictive performance of models utilizing exclusively technical variables against hybrid model configurations.

Given the binary nature of the target variable \( Y(n) \), model performance was assessed using a comprehensive suite of classification metrics. The evaluation criteria included:
\begin{itemize}
    \item The Area Under the Receiver Operating Characteristic Curve (AUC) [\cite{hanley1982meaning}].
    \item Accuracy (ACC) [\cite{huang2005using}].
    \item Recall (Sensitivity) [\cite{davis2006relationship}].
    \item Analysis of Confusion Matrices.
    \item Lift Curves [\cite{gong2021novel}].
\end{itemize}

These metrics were computed for both training and testing subsets to monitor model behavior across different data partitions. To specifically quantify and address potential overfitting, we introduced two differential metrics:
\begin{eqnarray}
\text{AUC}_{\text{diff}} &=& \text{AUC}_{\text{train}} - \text{AUC}_{\text{test}} \label{eq:auc_diff}\\
\text{ACC}_{\text{diff}} &=& \text{ACC}_{\text{train}} - \text{ACC}_{\text{test}} \label{eq:acc_diff}
\end{eqnarray}
Smaller values for these differential metrics indicate better model generalization, reflecting minimized performance disparity between training and testing phases.

%%%%%%%%%%%%%%%%%%%%%%%%%%%%%%%%%%%%%%%%%%%%%%%%%%

\subsection{Model Configurations}
\label{s:methods.1.1}

Identifying the optimal combination of input variables and hyperparameters for a predictive model constitutes a complex optimization challenge in neural networks. To address this, we implemented a randomized search strategy, evaluating various variable sets and LSTM architectural configurations.

Ten distinct variable sets were constructed and evaluated, as detailed below.

\begin{itemize}
    \item \textbf{Model 0:} Price data only (baseline).
    \item \textbf{Model 1:} Technical indicators and oscillators.
    \item \textbf{Model 2:} Fundamental data exclusively.
    \item \textbf{Model 3:} Technical indicators, oscillators, and fundamental data.
    \item \textbf{Model 4:} Technical indicators, oscillators, and support/resistance levels.
    \item \textbf{Model 5:} Technical indicators, oscillators, support/resistance levels, and fundamental data.
    \item \textbf{Model 6:} Technical indicators, oscillators, support/resistance levels, and divergence signals.
    \item \textbf{Model 7:} Technical indicators, oscillators, support/resistance levels, divergence signals, and fundamental data.
    \item \textbf{Model 8:} Technical indicators, oscillators, support/resistance levels, divergence signals, and Fibonacci levels.
    \item \textbf{Model 9:} All available features: technical indicators, oscillators, support/resistance levels, divergence signals, Fibonacci levels, and fundamental data.
\end{itemize}

For each variable set, 18 distinct model instances were trained by systematically exploring different hyperparameter combinations. This resulted in a total of $10 \times 18 = 180$ unique model configurations being evaluated. Performance was assessed using AUC and Accuracy metrics on both training and test partitions.

The hyperparameter search space was defined as follows:
\begin{itemize}
    \item \textbf{Training Epochs:} 20, 40.
    \item \textbf{LSTM Layers:} 1, 4, 8.  
    \item \textbf{Look-back Window:} 20, 30 days.
\end{itemize}

%%%%%%%%%%%%%%%%%%%%%%%%%%%%%%%%%%%%%%%%%%%%%%

\subsection{Results}
\label{s:methods.1.2}

The challenge of identifying the optimal LSTM configuration and variable set was formulated as an optimization problem, where (i) the inputs comprise the feature sets and LSTM hyperparameters, and (ii) the output is the set of machine learning metrics with the highest predictive capability.

Given the complexity of multi-metric analysis, AUC was selected as the primary performance measure. For each model configuration, we define AUC\textsubscript{min} as the minimum value between the training and test AUC across all 18 hyperparameter implementations  of that model:
\begin{equation}
\text{AUC\textsubscript{min}} = \min \{\text{AUC\textsubscript{train}}, \text{AUC\textsubscript{test}}\} \label{eq:auc_min}  
\end{equation}

This conservative metric ensures robustness against overfitting. Additionally, overfitting was quantified using the AUC\textsubscript{diff} metric defined in Equation (\ref{eq:auc_diff}).

We first present aggregated results across the 18 hyperparameter combinations for each model (Models 0-9). For each model, we computed the maximum, average, and minimum of AUC\textsubscript{min}, along with the average AUC\textsubscript{diff}. The results are summarized in Table \ref{tab:results_auc_per_model}.

\begin{table}[ht]
\centering
\begin{tabular}{|c|c|c|c|c|}
\hline
Model & MAX(AUC\textsubscript{min}) & AVG(AUC\textsubscript{min}) & MIN(AUC\textsubscript{min}) & AVG(AUC\textsubscript{diff}) \\ \hline
Model 0 & 0.56 & 0.55 & 0.55 & 0.06 \\ 
Model 1 & 0.57 & 0.48 & 0.42 & 0.13 \\
Model 2 & 0.65 & 0.59 & 0.43 & 0.04 \\ 
Model 3 & 0.64 & 0.57 & 0.51 & 0.15 \\ 
Model 4 & 0.54 & 0.47 & 0.43 & 0.12 \\ 
Model 5 & 0.64 & 0.58 & 0.53 & 0.16 \\ 
Model 6 & 0.53 & 0.48 & 0.40 & 0.12 \\ 
Model 7 & 0.64 & 0.57 & 0.53 & 0.20 \\ 
Model 8 & 0.53 & 0.47 & 0.43 & 0.12 \\ 
Model 9 & 0.62 & 0.57 & 0.51 & 0.19 \\ \hline
\end{tabular}
\caption{Aggregated AUC results across 18 hyperparameter configurations per model.}
\label{tab:results_auc_per_model}
\end{table}

The same aggregation was performed across different LSTM hyperparameter dimensions. Results by training epochs, network layers, and look-back window size are shown in Tables \ref{tab:results_auc_per_epochs}, \ref{tab:results_auc_per_layers}, and \ref{tab:auc_results_back_days}, respectively.

\begin{table}[ht]
\centering
\begin{tabular}{|c|c|c|c|c|}
\hline
Epochs & MAX(AUC\textsubscript{min}) & AVG(AUC\textsubscript{min}) & MIN(AUC\textsubscript{min}) & AVG(AUC\textsubscript{diff}) \\ \hline
20 & 0.64 & 0.53 & 0.42 & 0.09 \\
40 & 0.64 & 0.53 & 0.43 & 0.13 \\
60 & 0.65 & 0.53 & 0.40 & 0.17 \\ \hline
\end{tabular}
\caption{Aggregated AUC results by number of training epochs.}
\label{tab:results_auc_per_epochs}
\end{table}

\begin{table}[ht]
\centering
\begin{tabular}{|c|c|c|c|c|}
\hline
Layers & MAX(AUC\textsubscript{min}) & AVG(AUC\textsubscript{min}) & MIN(AUC\textsubscript{min}) & AVG(AUC\textsubscript{diff}) \\ \hline
1 & 0.62 & 0.51 & 0.40 & 0.18 \\
4 & 0.65 & 0.55 & 0.44 & 0.13 \\
8 & 0.61 & 0.53 & 0.43 & 0.08 \\ \hline
\end{tabular}
\caption{Aggregated AUC results by number of LSTM layers.}
\label{tab:results_auc_per_layers}
\end{table}

\begin{table}[ht]
\centering
\begin{tabular}{|c|c|c|c|c|}
\hline
Back Days & MAX(AUC\textsubscript{min}) & AVG(AUC\textsubscript{min}) & MIN(AUC\textsubscript{min}) & AVG(AUC\textsubscript{diff}) \\ \hline
20 & 0.65 & 0.53 & 0.42 & 0.087 \\
30 & 0.64 & 0.54 & 0.40 & 0.127 \\ \hline
\end{tabular}
\caption{AUC results by look-back window size.}
\label{tab:auc_results_back_days}
\end{table}

Analysis reveals that models incorporating fundamental data (Models 2, 3, 5, 7, and 9) consistently achieve superior AUC metrics across aggregation types. However, overfitting tends to increase with model complexity, as evidenced by rising AUC\textsubscript{diff} values.

Regarding optimal configurations, Tables \ref{tab:results_auc_per_epochs} and \ref{tab:results_auc_per_layers} indicate that overfitting increases with training epochs but decreases with additional layers, reaching minimal levels in 8-layer architectures. Table \ref{tab:auc_results_back_days} shows modest performance differences between 20 and 30-day look-back windows.

This preliminary analysis suggests that optimal performance is achieved among Models 2, 3, 5, 7, and 9, with LSTM configurations of 4 layers, 20-40 epochs, and 20-30 day look-back windows.

From this candidate set, we performed a refined selection based on AUC\textsubscript{min} maximization. Table \ref{tab:model_best_performance} presents the top-performing configurations for each selected model.

\begin{table}[ht]
\centering
\begin{tabular}{|c|c|c|c|c|c|c|c|c|}
\hline
Model & Features & Epochs & Layers & Days & AUC\textsubscript{test} & AUC\textsubscript{train} & AUC\textsubscript{min} & AUC\textsubscript{diff} \\ \hline
2 & 39 & 60 & 4 & 20 & 0.65 & 0.68 & 0.65 & 0.03 \\
3 & 83 & 20 & 4 & 30 & 0.64 & 0.67 & 0.64 & 0.02 \\
7 & 91 & 20 & 4 & 20 & 0.66 & 0.64 & 0.64 & 0.02 \\
5 & 87 & 40 & 4 & 20 & 0.64 & 0.76 & 0.64 & 0.12 \\
9 & 95 & 20 & 4 & 20 & 0.62 & 0.65 & 0.62 & 0.02 \\ \hline
\end{tabular}
\caption{Optimal configurations for top-performing models based on AUC\textsubscript{min} maximization.}
\label{tab:model_best_performance}
\end{table}

The top three models demonstrate promising metrics, with AUC\textsubscript{min} values of 0.65-0.64 and minimal overfitting (AUC\textsubscript{diff} = 0.02-0.03). Notably, Model 2 achieves optimal performance with 60 epochs, extending beyond the initially specified search range of 20-40 epochs. This suggests that the preliminary analysis served to identify promising regions of the hyperparameter space, while detailed configuration optimization revealed additional performance gains at extended training durations.

Given the comparable performance metrics among the top three models, we proceeded to trading strategy simulations to evaluate their behavior with real market data.

%%%%%%%%%%%%%%%%%%%%%%%%%%%%%%%%%%%%%%%

\section{Trading Simulation and Performance Evaluation}
\label{s:application}

Following the evaluation of machine learning metrics, we conducted comprehensive trading simulations to assess the practical predictive capability and financial viability of the selected models. These simulations utilized out-of-sample data to ensure robust performance validation.

\subsection{Simulation Framework}

The testing period spanned from June 1, 2023, to March 4, 2024. For each trading day, the three top-performing models (Models 2, 3, and 7) generated probability forecasts. To address the inherent bias in probability calibration due to imbalanced data distributions, we applied min-max normalization to obtain weighted probabilities:

\begin{equation}
\mathrm{Prob}_{\text{weighted}} = \frac{{\mathrm{Prob} - \mathrm{Prob}_{\text{min}}}}{{\mathrm{Prob}_{\text{max}} - \mathrm{Prob}_{\text{min}}}}
\end{equation}
where $\mathrm{Prob}_{\text{weighted}}$ is the normalized probability score, $\mathrm{Prob}$ is the raw model output, and $\mathrm{Prob}_{\text{min}}$, $\mathrm{Prob}_{\text{max}}$ are the minimum and maximum probabilities observed in the test predictions, respectively.

Trading thresholds were established at 0.7 for long positions and 0.35 for short positions, with the latter adjusted from an initial 0.3 threshold due to the absence of signals below 0.35 in preliminary testing.

\subsection{Fixed-Horizon Trading Simulation}

The first simulation implemented a fixed-horizon strategy where positions were automatically closed after 10 days, regardless of market conditions. This approach facilitated statistical validation of model predictive capabilities. Results are summarized in Table \ref{tab:app1_trades_summary}.

\begin{table}[ht]
\centering
\begin{tabular}{|l|l|c|c|c|c|}
\hline
\textbf{Model} & \textbf{Position} & \textbf{Winning Trades} & \textbf{Losing Trades} & \textbf{Total Return} & \textbf{Win Rate (\%)} \\ \hline
Model 2 & Long & 33 & 51 & -7.55\% & 39.29 \\ \hline
Model 2 & Short & 9 & 2 & 20.53\% & 81.82 \\ \hline
Model 3 & Long & 84 & 106 & -1.23\% & 44.21 \\ \hline
Model 3 & Short & 9 & 2 & 73.17\% & 81.82 \\ \hline
Model 7 & Long & 12 & 2 & 69.64\% & 85.71 \\ \hline
Model 7 & Short & 6 & 0 & 80.27\% & 100.00 \\ \hline
\end{tabular}
\caption{Fixed-horizon trading performance (10-day holding period)}
\label{tab:app1_trades_summary}
\end{table}

Notably, Model 7 demonstrated exceptional performance across both long and short positions, achieving the highest win rates and substantial positive returns. All models exhibited superior capability in predicting downward movements (short positions), suggesting enhanced sensitivity to EUR/USD decline signals.

\subsection{Dynamic Position Management Simulation}

The second simulation employed a dynamic strategy where positions remained open as long as the model continued to generate signals above the established thresholds. This approach more closely resembles real-world trading by minimizing transaction costs and avoiding premature position closure. Results are presented in Table \ref{tab:app2_trades_summary}.

\begin{table}[ht]
\centering
\begin{tabular}{|l|l|l|l|l|l|l|}
\hline
\textbf{Model} & \textbf{Position} & \textbf{Entry Date} & \textbf{Exit Date} & \textbf{Entry Price} & \textbf{Exit Price} & \textbf{Return (\%)} \\ \hline
Model 2 & Long & 2023-07-18 & 2023-08-15 & 1.123760 & 1.090417 & -2.97 \\ \hline
Model 2 & Long & 2023-08-14 & 2023-09-23 & 1.094439 & 1.066155 & -2.58 \\ \hline
Model 2 & Long & 2023-10-14 & 2023-11-26 & 1.053674 & 1.090631 & 3.51 \\ \hline
Model 2 & Short & 2023-06-21 & 2023-07-11 & 1.092037 & 1.100594 & -0.78 \\ \hline
Model 3 & Long & 2023-08-18 & 2024-03-03 & 1.087465 & 1.080497 & -0.64 \\ \hline
Model 3 & Long & 2024-02-24 & 2024-03-05 & 1.082567 & 1.085305 & 0.25 \\ \hline
Model 3 & Short & 2023-07-22 & 2023-08-11 & 1.113710 & 1.098165 & 1.40 \\ \hline
Model 7 & Long & 2023-11-05 & 2023-11-18 & 1.061909 & 1.085376 & 2.21 \\ \hline
Model 7 & Long & 2024-02-14 & 2024-03-03 & 1.070893 & 1.080497 & 0.90 \\ \hline
Model 7 & Long & 2024-02-24 & 2024-03-05 & 1.082567 & 1.085305 & 0.25 \\ \hline
Model 7 & Short & 2023-08-13 & 2023-08-28 & 1.098165 & 1.086921 & 1.02 \\ \hline
\end{tabular}
\caption{Dynamic position management trading performance}
\label{tab:app2_trades_summary}
\end{table}

The dynamic approach significantly reduced trading frequency while maintaining profitability. Model 7 achieved perfect performance across all four executed trades (three long, one short), reinforcing its superiority identified in machine learning metrics. While Models 2 and 3 showed limited success in long positions, their short-position performance remained competitive.

\subsection{Transaction Cost Analysis}

The practical implementation of any trading strategy must account for transaction costs, which can substantially impact net returns. In forex markets, several cost components require consideration:

\begin{enumerate}
    \item \textbf{Spread}: The bid-ask differential represents a primary cost component. Variable spreads typically range from 0.1 to 3 pips for major currency pairs like EUR/USD during normal market conditions.
    
    \item \textbf{Commissions}: ECN brokers typically charge commissions per traded lot, often ranging from \$2-\$7 per round turn for standard lots.
    
    \item \textbf{Swap Rates}: Overnight financing costs can accumulate for positions held beyond daily settlement, calculated from interbank interest rate differentials.
    
    \item \textbf{Slippage}: Execution variance during high volatility periods may impact entry and exit prices, particularly for larger position sizes.
    
    \item \textbf{Ancillary Fees}: Account maintenance, deposit/withdrawal, and inactivity fees may apply depending on broker policies.
\end{enumerate}

These cost factors collectively influence strategy profitability and should be incorporated into comprehensive performance evaluation [\cite{hennart2005transaction}]. For our simulations, we assumed average spread costs of 1 pip for EUR/USD and commission-free trading, representing competitive retail trading conditions.

\section{Discussion}
\label{s:discussion}

The development of a robust algorithmic trading system presents considerable complexity, encompassing multiple interdependent optimization challenges. This work addressed the dual problem of (i) selecting optimal feature sets and (ii) optimizing LSTM network hyperparameters for each configuration. The inherent difficulty stems from the multi-objective nature of model optimization, where improvements in one metric often come at the expense of others. For instance, maximizing Accuracy may compromise AUC performance, while controlling overfitting through regularization can reduce predictive accuracy. Furthermore, architectural decisions such as increasing network depth necessitate longer training periods, thereby elevating overfitting risks.

Given these competing objectives and the high-dimensional parameter space, identifying a global optimum proved computationally intractable. Consequently, our methodology focused on identifying strong local optima by prioritizing AUC maximization while constraining overfitting through the AUC\textsubscript{diff} metric. This approach acknowledges the practical trade-offs inherent in real-world trading system development. Additionally, the stochastic nature of neural network parameter initialization introduces further variability that must be considered when evaluating model performance.

The validation pipeline extended beyond conventional machine learning metrics to include realistic trading simulations with out-of-sample data. This comprehensive evaluation required careful calibration of trading thresholds and position management rules to translate probabilistic forecasts into executable strategies. The simulation results demonstrate that despite the optimization challenges, it is computationally feasible to emulate sophisticated trading decision-making processes.

Notably, our findings reveal that fundamental data exhibited significantly greater predictive power than technical indicators alone. Models incorporating fundamental variables (particularly Models 2, 3, 5, 7, and 9) consistently outperformed those relying exclusively on technical analysis. This suggests that macroeconomic factors provide valuable signals that complement price-based technical indicators in forecasting EUR/USD movements.

Most significantly, the results indicate that carefully optimized neural network architectures can achieve sufficient statistical edge to develop profitable trading systems, even in highly efficient markets like Forex. The success of Model 7 across multiple evaluation frameworks—from machine learning metrics to both fixed-horizon and dynamic trading simulations—underscores the potential of systematic approaches to currency market prediction.

\section{Conclusion}
\label{s:conclusion}

This study confirms the central hypothesis that integrating both technical and fundamental variables within a unified prediction framework significantly enhances forecasting capability in algorithmic trading systems. We have demonstrated the feasibility of constructing a comprehensive cognitive system capable of computationally processing the diverse information sets that human traders typically analyze. The implemented models successfully incorporated multiple data categories, including fundamental indicators, technical oscillators, support and resistance levels, Fibonacci retracements, and divergence signals.

The machine learning metrics achieved by our optimized models surpassed performance levels reported in prior literature, suggesting the potential for obtaining a statistical advantage in market forecasting. Crucially, we extended beyond conventional model evaluation by implementing realistic trading simulations that validated the practical applicability of our approach in near-real-world conditions.

Regarding the LSTM architecture optimization, while superior configurations may exist within the vast hyperparameter space, our identified setup—comprising 4 layers, 20 training epochs, and a 20-day lookback window, with L1 dropout regularization of 0.1—demonstrated robust performance. The selection of standard values for additional parameters (learning rate, decay, momentum) represented a pragmatic compromise given computational constraints and the combinatorial complexity of exhaustive search.

Feature engineering emerged as the most critical and resource-intensive aspect of model development. The meticulous selection and construction of both technical and fundamental variables proved essential to model performance. Although the computational cost of calculating dynamic support and resistance levels was substantial, their inclusion yielded positive results. Conversely, Fibonacci levels did not contribute meaningfully to predictive accuracy, potentially due to information redundancy or the network's inability to extract relevant patterns given the chosen architecture.

The trading simulation implementing dynamic position management closely approximated real-world conditions and represents a critical preliminary step toward live deployment. The necessity to manually calibrate the short-position threshold (adjusted from 0.3 to 0.35) highlights an important limitation and suggests future research directions in adaptive threshold optimization.

Several promising avenues for further investigation emerge from this work. Future research could explore alternative target variable formulations, multi-time frame integration (minute, hourly, weekly), incorporation of geopolitical risk factors, and replication across additional currency pairs. Enhanced hyperparameter optimization techniques and automated threshold calibration methods represent additional productive directions.

In conclusion, this work provides affirmative evidence that algorithmic trading systems can surpass human decision-making capabilities in currency markets. While human traders face cognitive limitations in simultaneously processing numerous variables, computational systems offer virtually unlimited analytical capacity. However, this advantage comes with significant complexity—the development process involves numerous interdependent parameters where singular misconfigurations can substantially impact performance. The demonstrated framework establishes a foundation for continued advancement in systematic trading approaches that leverage diverse data sources through sophisticated machine learning architectures.


\begin{thebibliography}{99}

\bibitem{alostad2017directional} Alostad H.; Davulcu, H. Directional prediction of stock prices using breaking news on Twitter, 2015. IEEE/WIC/ACM International Conference on Web Intelligence and Intelligent Agent Technology (WI-IAT), Singapore, 2015, pp. 523-530.


\bibitem{anuar2025comparative} Anuar, A. A., A. A. B. Sulaiman, and M. T. B. Mohamad. Comparative analysis of AI-driven versus human-managed equity funds across market trends. \textit{Future Business Journal} \textbf{2025}, 11(1), 95.

\bibitem{appel1985moving} Appel, G. \textit{The Moving Average Convergence-Divergence Trading Method: Advanced Version}. Scientific Investment Systems, 1985.

\bibitem{blau1995momentum} Blau, W. \textit{Momentum, Direction, and Divergence}. John Wiley \& Sons, 1995.

\bibitem{bollinger2002bollinger} Bollinger, J. \textit{Bollinger on Bollinger Bands}. McGraw-Hill, New York, 2002.

\bibitem{chen2010news} Chen, Y.-L. and Y.-F. Gau. News announcements and price discovery in foreign exchange spot and futures markets. \textit{Journal of Banking \& Finance} \textbf{2010}, 34(7), 1628–1636.

\bibitem{cohen2022algorithmic} Cohen, G. Algorithmic trading and financial forecasting using advanced artificial intelligence methodologies. \textit{Mathematics} \textbf{2022}, 10(18), 3302.

\bibitem{davis2006relationship} Davis, J. and M. Goadrich. The relationship between precision-recall and ROC curves. In \textit{Proceedings of the 23rd International Conference on Machine Learning}, 2006, pp. 233–240.

\bibitem{demirtacs2020algorithmic} Demirtaş, Ş. C. and S. Ç. Şahin. Algorithmic trading versus human traders at different information levels. \textit{CEEOL} \textbf{2020}, 1, 1–13.

\bibitem{elliott2007ichimoku} Elliott, N. \textit{Ichimoku Charts: An Introduction to Ichimoku Kinko Clouds}. Harriman House Limited, 2007.

\bibitem{fischer2018deep} Fischer, T. and C. Krauss. Deep learning with long short-term memory networks for financial market predictions. \textit{European Journal of Operational Research} \textbf{2018}, 270(2), 654–669.

\bibitem{gong2021novel} Gong, M. A novel performance measure for machine learning classification. \textit{International Journal of Managing Information Technology (IJMIT)} \textbf{2021}, 13, 1–5.

\bibitem{hanley1982meaning} Hanley, J. A. and B. J. McNeil. The meaning and use of the area under a receiver operating characteristic (ROC) curve. \textit{Radiology} \textbf{1982}, 143(1), 29–36.

\bibitem{hennart2005transaction} Hennart, J.-F. Transaction costs theory and the multinational enterprise. \textit{The Nature of the Transnational Firm} \textbf{2005}, 1(1), 79–126.

\bibitem{huang2005using} Huang, J. and C. X. Ling. Using AUC and accuracy in evaluating learning algorithms. \textit{IEEE Transactions on Knowledge and Data Engineering} \textbf{2005}, 17(3), 299–310.

\bibitem{jacob2024algorithmic} Jacob-Leal, S. and N. Hanaki. Algorithmic trading, what if it is just an illusion? Evidence from experimental asset markets. \textit{Journal of Behavioral and Experimental Economics} \textbf{2024}, 112, 102240.

\bibitem{king2024blockchain} King, J.C.; Dale, R.; Amigó, J.M. Blockchain metrics and indicators in cryptocurrency trading. Chaos, Solitons \& Fractals \textbf{2024}, 178, 114305.


\bibitem{king2025integration} King, J. C. and J. M. Amigó. Integration of LSTM networks in random forest algorithms for stock market trading predictions. \textit{Forecasting} \textbf{2025}, 7(3), 49.


\bibitem{kirkpatrick2010technical} Kirkpatrick, C. D. and J. R. Dahlquist. \textit{Technical Analysis: The Complete Resource for Financial Market Technicians}. FT Press, Upper Saddle River, NJ, 2010.

\bibitem{klinker2011exponential} Klinker, F. Exponential moving average versus moving exponential average. \textit{Mathematische Semesterberichte} \textbf{2011}, 58, 97–107.

\bibitem{lecun2015deep} LeCun, Y., Y. Bengio, and G. Hinton. Deep learning. \textit{Nature} \textbf{2015}, 521(7553), 436–444.

\bibitem{li2020intelligent} Li, X. and C. Luo. An intelligent stock trading decision support system based on rough cognitive reasoning. \textit{Expert Systems with Applications} \textbf{2020}, 160, 113763.

\bibitem{liaudinskas2022human} Liaudinskas, K. Human vs. machine: Disposition effect among algorithmic and human day-traders. \textit{Norges Bank Working Paper} \textbf{2022}, 6/2022.

\bibitem{martin2019cognitive} Martín Parrondo, R. Cognitive trading system model. \textit{SSRN Electronic Journal} \textbf{2019}, 1, 1–13.

\bibitem{mazya2018approximate} Maz’ya, V. and G. Schmidt. \textit{Approximate Approximations}. American Mathematical Society, 2018.

\bibitem{murphy1999technical} Murphy, J. J. \textit{Technical Analysis of the Financial Markets: A Comprehensive Guide to Trading Methods and Applications}. New York Institute of Finance, 1999.

\bibitem{osler2000support} Osler, C. L. Support for resistance: Technical analysis and intraday exchange rates. \textit{Economic Policy Review} \textbf{2000}, 6(2), 54–68.

\bibitem{panapongpakorn2019short} Panapongpakorn, T. and D. Banjerdpongchai. Short-term load forecast for energy management systems using time series analysis and neural network method with average true range. In \textit{Proceedings of the 2019 International Symposium on Instrumentation, Control, Artificial Intelligence, and Robotics (ICA-SYMP)}, IEEE, pp. 86–89.

\bibitem{reznik2018high} Reznik, N. and L. Pankratova. High-frequency trade as a component of algorithmic trading: Market consequences. In \textit{ICTERI Workshops}, 2018, pp. 73–83.


\bibitem{salkar2021algorithmic} Salkar, T.; Shinde, A.; Tamhankar, N.; Bhagat, N. Algorithmic trading using technical indicators. In 2021 International Conference on Communication information and Computing Technology (ICCICT), 2021, pp. 1--6.


\bibitem{shaker2018predictive} Shaker, R. Z., M. Asad, and N. Zulfiqar. Do predictive power of Fibonacci retracements help the investor to predict future? A study of Pakistan Stock Exchange. \textit{International Journal of Economics and Financial Research} \textbf{2018}, 4(6), 159–164.

\bibitem{steele2015technical} Steele, R.; Esmahi, L. Technical indicators as predictors of position outcome for technical 665 trading. In Proceedings of the International Conference on e-Learning, e-Business, Enterprise Information Systems, and e-Government (EEE), 2015, p. 3.


\bibitem{thompson1947use} Thompson, W. R. Use of moving averages and interpolation to estimate median effective dose: I. Fundamental formulas, estimation of error, and relation to other methods. \textit{Bacteriological Reviews} \textbf{1947}, 11(2), 115–145.

\bibitem{wang2022cognitive} Wang, Y., L. Zhang, and M. Li. Cognitive computing: Definition, technologies, and research issues. \textit{Cognitive Computation} \textbf{2022}, 14(2), 456–472.

\bibitem{wu2015technical} Wu, M.; Diao, X. Technical analysis of three stock oscillators testing macd, rsi and kdj rules in sh \& sz stock markets. In 2015 4th International Conference on Computer Science and Network Technology (ICCSNT), 2015, Volume 1, pp. 320--323.


\bibitem{wu2015oscillators} Wu, M. and X. Diao. Technical analysis of three stock oscillators: Testing MACD, RSI and KDJ rules in SH \& SZ stock markets. In \textit{2015 4th International Conference on Computer Science and Network Technology (ICCSNT)}, IEEE, pp. 320–323.

\bibitem{Zalinescu02} C. Z\u{a}linezcu. Convex Analysis in General Vector Spaces. World Scientific Publishing, Singapore, 2002.


\end{thebibliography}
\end{document}